\pdfoutput=1
\documentclass[10pt,twocolumn,letterpaper]{article}

\usepackage[pagenumbers]{cvpr}

\usepackage{pifont}      %
\usepackage{multirow}
\usepackage{makecell}
\usepackage{colortbl}    %
\definecolor{ourrow}{RGB}{228,239,250}
\makeatletter
\newcommand{\shadeL}{\gdef\CT@cell@color{\@tempdimb\z@\global\let\CT@cell@color\relax}}
\newcommand{\shadeR}{\gdef\CT@cell@color{\@tempdimc\z@\global\let\CT@cell@color\relax}}
\makeatother
\newcommand{\cmark}{\ding{51}}
\newcommand{\xmark}{\ding{55}}
\newcommand{\SN}{\mathcal{SN}}
\newcommand{\Vth}{V_{\mathrm{th}}}
\newcommand{\Din}{D_{\mathrm{in}}}
\newcommand{\Esop}{E_{\mathrm{SOP}}}
\newcommand{\Emac}{E_{\mathrm{MAC}}}

\definecolor{cvprblue}{rgb}{0.21,0.49,0.74}
\usepackage[pagebackref,breaklinks,colorlinks,allcolors=cvprblue]{hyperref}

\title{Spiking World-Action Models for Robotic Manipulation}

\author{
Mingquan Liu\textsuperscript{1,2}\quad
Peng Zhou\textsuperscript{2,}\thanks{Corresponding authors: Malu Zhang (\href{mailto:maluzhang@uestc.edu.cn}{maluzhang@uestc.edu.cn}) and Peng Zhou (\href{mailto:zp.pengzhou@gmail.com}{zp.pengzhou@gmail.com}).}\quad
Shanming Yao\textsuperscript{2,3}\quad
Shuai Wang\textsuperscript{1}\quad
Simo Wu\textsuperscript{2}\quad
Malu Zhang\textsuperscript{1,}\footnotemark[2]\\[0.4em]
\textsuperscript{1}University of Electronic Science and Technology of China\qquad
\textsuperscript{2}LuxiTech\\
\textsuperscript{3}Northwestern Polytechnical University
}
\hypersetup{pdftitle={Spiking World-Action Models for Robotic Manipulation},
  pdfauthor={Mingquan Liu, Peng Zhou, Shanming Yao, Shuai Wang, Simo Wu, Malu Zhang}}

\begin{document}
\setcounter{footnote}{1}%
\maketitle
\begin{abstract}
World action models (WAMs) predict how the world responds to actions and offer a promising path toward general embodied intelligence, but they are large and run the network many times per plan, so they are hard to deploy on robots.
Spiking neural networks (SNNs) compute with sparse binary spikes and use additions instead of multiplications, which saves power.
However, existing WAMs use dense videos as the world state, so a spiking version would spend most of its computation on the background.
We therefore introduce SpikingWAM, the first spiking WAM, which uses a sparse task skeleton as the world state.
Inspired by biological motor control, which attends only to the key positions of a task, the Task Skeleton State (TSS) keeps the keypoints of the arm and the task objects, and a small spiking front end estimates the object keypoints from images.
Body Frame Spike Coding (BFSC) and Grid Coincidence Attention (GCA) keep millimetre precision, and the Persistent Single Step Readout (PSR) runs the network once per plan.
On the simulated benchmarks LIBERO, LIBERO-Plus and RoboTwin 2.0, SpikingWAM achieves competitive success rates with far less energy than ANN methods and surpasses all compared methods on LIBERO-Plus.
Real robot experiments further confirm its effectiveness.
These results show that a task skeleton lets spiking networks implement WAMs at very low energy, offering a new solution for low power embodied intelligence.
Project page: \href{https://wxqnl.github.io/spikingwam/}{https://wxqnl.github.io/spikingwam}
\end{abstract}
\begin{figure}[t]
  \centering
  \includegraphics[width=\linewidth]{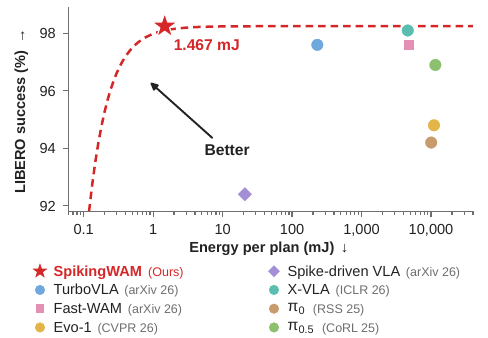}
  \caption{\textbf{Energy and success of SpikingWAM and other VLAs and WAMs on LIBERO. The upper left is better.}
  Energy is the estimate per plan for a 45\,nm process (\cref{sec:setup}).
  The dashed line is a dividing line.}
  \label{fig:teaser}
\end{figure}

\section{Introduction}
\label{sec:intro}

World action models (WAMs)~\cite{dreamzero,fastwam,kim2026cosmos,sheng2026skelewam,geact} let a robot learn in one model how to act and how the world changes in response.
They offer a promising path toward general embodied intelligence.
However, most existing WAMs use videos or visual latents as the world state~\cite{dreamzero,fastwam,kim2026cosmos,geact}.
These models are large and hard to run on robots with limited resources.
To this end, recent work skips future generation at inference~\cite{fastwam,sheng2026skelewam}, which reduces computation.
These models still rely on dense multiply and accumulate operations, and each plan runs the network several times for multistep sampling.
A robot in closed loop control plans again and again, so this cost adds up.
Lowering the energy of each plan is therefore key to deploying WAMs on board.

Neurons in spiking neural networks (SNNs) communicate with discrete spikes and fire only when the membrane potential crosses a threshold.
On neuromorphic hardware, this event-driven computation replaces multiply and accumulate with accumulate, so its energy grows with the number of events~\cite{roy2019towards}.
SNNs have recently moved from visual perception~\cite{yao2025scaling} to embodied control.
Spiking vision-language-action models (VLAs)~\cite{song2026spikevla,wang2026spikedriven} and spiking diffusion policies~\cite{hou2024sdp,wang2024stmdp} generate actions, and a spiking world model~\cite{sun2025spiking} predicts the next state after an action.
However, WAMs are hard to implement directly with SNNs.
Mainstream WAMs are built on large video generation models~\cite{dreamzero,fastwam,kim2026cosmos,geact}, and there is still no mature way to turn such models into SNNs.
Even if this were done, dense video representations would still spend most events on a background unrelated to the task.
Small jitter in the image would also keep triggering new events.
Running the network several times per plan further cancels the savings from sparsity.
Existing approaches therefore cannot exploit the sparse and event-driven nature of SNNs.
Integer spikes also struggle to keep the millimetre precision of fine manipulation.

In contrast, biological motor control attends to the key positions of a task and represents space relative to the body.
When people manipulate objects, their gaze falls almost only on grasp points and targets~\cite{johansson2001eye}.
The premotor and posterior parietal cortices represent targets relative to the arm or the hand~\cite{graziano1994coding,buneo2002direct}.
Grid cells encode position with periodic firing~\cite{hafting2005microstructure}, and cortical neurons can detect spikes that arrive at the same time~\cite{konig1996integrator}.
This suggests that the world state only needs the key positions of the arm and the task objects.
Once coordinates are written into spikes relative to the body, the positions of hand and object can be compared by spike coincidence.
Following the event-driven principle, only changes pass between consecutive plans.

Based on these ideas, we propose a sparse and event-driven paradigm for spiking WAMs and build SpikingWAM on it.
To our knowledge, SpikingWAM is the first spiking world action model.
A single spiking network trained from scratch learns future skeletons and action chunks together, and it generates only the action chunks at inference (\cref{fig:arch}).
First, we propose the Task Skeleton State (TSS), which uses only the keypoints of the arm and the task objects as the world state.
A small spiking perception front end estimates the object keypoints from images.
Next, we propose Body Frame Spike Coding (BFSC) and Grid Coincidence Attention (GCA).
BFSC encodes coordinates relative to the body and GCA compares positions by spike coincidence, so the geometry keeps millimetre precision.
Finally, we propose the Persistent Single Step Readout (PSR).
It runs the network once per plan and passes only the changes between consecutive plans.
On LIBERO~\cite{liu2023libero} and LIBERO-Plus~\cite{fei2026liberoplus}, SpikingWAM reaches success rates comparable to state-of-the-art (SOTA) artificial neural network (ANN) methods.
On bimanual manipulation in RoboTwin 2.0~\cite{chen2025robotwin}, it reaches a mean success rate of 80.20\% without pretraining on robot data.
On LIBERO, it needs only 1.467\,mJ per plan including the perception front end, which is more than an order of magnitude below every compared method (\cref{fig:teaser}).
On a real robot arm, SpikingWAM completes a pick and place task given by a language instruction, with a success rate comparable to ANN VLAs and only 3.07\,mJ per plan.
Our main contributions are as follows.
\begin{itemize}
  \item We design TSS and a small spiking perception front end that estimates its object keypoints from images.
  They focus the events and attention of the policy on the parts that matter for the task, which lowers the energy of each plan.
  \item We propose BFSC and GCA to recover the millimetre precision that integer spikes lose.
  We propose PSR and an offline check that tells when it applies, so the network runs only once per plan.
  \item We propose a sparse and event-driven paradigm for spiking WAMs and build on it SpikingWAM, the first spiking world action model.
  It reaches 98.25\%, 88.76\% and 80.20\% on LIBERO, LIBERO-Plus and RoboTwin 2.0.
  On LIBERO-Plus, it exceeds all compared methods.
\end{itemize}
\section{Related Work}
\label{sec:related}

\subsection{World Action Models and Fast Sampling}

VLA models generate actions directly from camera images and language instructions~\cite{kim2024openvla,black2024pi0,intelligence2025pi05}, and compact designs further reduce the inference cost~\cite{xie2026turbovla}.
World action models build on this by adding future prediction, so that the model also learns how the scene evolves.
Mainstream approaches build on pretrained video models and generate future frames together with actions~\cite{dreamzero,kim2026cosmos,geact}, which makes them large.
Recent work steadily reduces the cost of the world state.
VLA-JEPA~\cite{sun2026vlajepa} replaces pixel reconstruction with visual latents.
Fast-WAM~\cite{fastwam} shows that the benefit of future prediction comes mainly from joint training, so future generation can be skipped at inference.
SkeleWAM~\cite{sheng2026skelewam} represents the world state as a 3D skeleton and compresses the scene into a few keypoints.
In parallel, policy distillation~\cite{prasad2024consistency,wang2024onedp} and mean flows~\cite{sheng2026mp1,geng2025meanflow} compress action generation into one or a few steps.
However, while these works shrink the model and the sampling steps, the computation itself still consists of dense multiply and accumulate operations.
Repeated planning in closed loop control thus still costs too much energy on board.

\subsection{SNNs and Spiking Embodied Models}

Spiking Transformers allow SNNs to handle vision tasks at a large scale.
Spiking self-attention~\cite{zhou2023spikformer}, spike-driven attention~\cite{yao2023spikedriven,yao2024spike} and integer training~\cite{yao2025scaling} have steadily improved their scale and accuracy.
This progress has carried SNNs into embodied control along two lines.
One line generates actions.
Spiking VLAs~\cite{song2026spikevla,wang2026spikedriven} and spiking diffusion policies~\cite{hou2024sdp,wang2024stmdp} have been applied to navigation and manipulation.
The other line predicts states.
A spiking world model~\cite{sun2025spiking} learns the next state after a given action for model-based reinforcement learning.
A separate network produces the actions.
However, the two lines have developed separately.
Action generation and scene prediction have not been unified in one spiking network, so world action models do not yet use event-driven computation.
\section{Preliminaries}
\label{sec:prelim}

\subsection{Spiking Neuron Model}
\label{sec:neuron}

A spiking neuron integrates its input into a membrane potential and fires a spike only when the potential crosses a threshold.
The synaptic input of layer $l$ is $\mathbf{I}^{l}=W^{l}\mathbf{s}^{l-1}$, where $W^{l}$ is the synaptic weight matrix and $\mathbf{s}^{l-1}$ holds the spikes fired by the previous layer.
A neuron that does not fire triggers no synaptic computation, so the next layer only accumulates the weights of the neurons that fire.
We use the bipolar integer spiking neuron~\cite{yao2025scaling}, written $\SN(\cdot)$,
\begin{align}
  \mathbf{s}^{l}&=\SN(\mathbf{u}^{l})=\mathrm{clamp}\big(\mathrm{round}(\mathbf{u}^{l}/\Vth^{l}),\,-D,\,D\big),\label{eq:neuron}\\
  \mathbf{s}^{l}&=\textstyle\sum_{d=1}^{D}\mathbf{s}^{l}[d],\qquad \mathbf{s}^{l}[d]\in\{-1,0,1\},\label{eq:unfold}
\end{align}
where $\mathbf{u}^{l}$ is the membrane potential, $\Vth^{l}$ is a learnable threshold, and $D$ is the largest number of spikes that a neuron fires in one forward pass.
During training the neuron outputs an integer.
At inference the integer output unfolds over $D$ time steps into unit spikes $\mathbf{s}^{l}[d]$ of the same sign.
The next layer then only adds or subtracts one weight column for each spike.
Within $[-D,D]$ the rounding passes the gradient with a straight through estimator, so the network can be trained end to end.

\subsection{Spiking Flow Matching}
\label{sec:sfm}

Flow matching~\cite{lipman2023flow,liu2023flow} learns a velocity field that transports Gaussian noise to the data distribution.
Given data $x$ and noise $\epsilon\sim\mathcal{N}(0,I)$, the straight path $x^{\tau}=(1-\tau)x+\tau\epsilon$ is pure noise at $\tau=1$.
The velocity field $v_{\theta}$ regresses the velocity along the path,
\begin{equation}
  \mathcal{L}_{\mathrm{FM}}(x)=\mathbb{E}_{\tau,\epsilon}\big\|v_{\theta}(x^{\tau},\tau\mid c)-(\epsilon-x)\big\|^{2},
  \label{eq:fm}
\end{equation}
where $c$ is the condition.
Sampling starts from $x^{1}=\epsilon$ and integrates to $\tau=0$ with $N$ Euler steps, $x^{\tau-1/N}=x^{\tau}-v_{\theta}(x^{\tau},\tau\mid c)/N$.
Spiking flow matching uses a spiking Transformer built from the neurons of \cref{eq:neuron} as $v_{\theta}$.
An input layer turns the noisy sample $x^{\tau}$ into membrane potentials.
After that, the weights of the self-attention and feedforward layers read only spikes.
Each integration step runs the spiking network once.

In a world action model, the condition $c$ is the language instruction $\ell$ and the current world state $o_t$.
The model applies flow matching to an action chunk $A_t$ of $H_a$ steps and to the future world state $O_t$ at the same time.
A readout with several samples~\cite{sheng2026skelewam} integrates $k$ action chunks and takes the medoid chunk, which is the sample with the smallest total distance to the others.
Each plan therefore runs the network $kN$ times.
\section{Method}
\label{sec:method}

\begin{figure}[t]
  \centering
  \includegraphics[width=\linewidth]{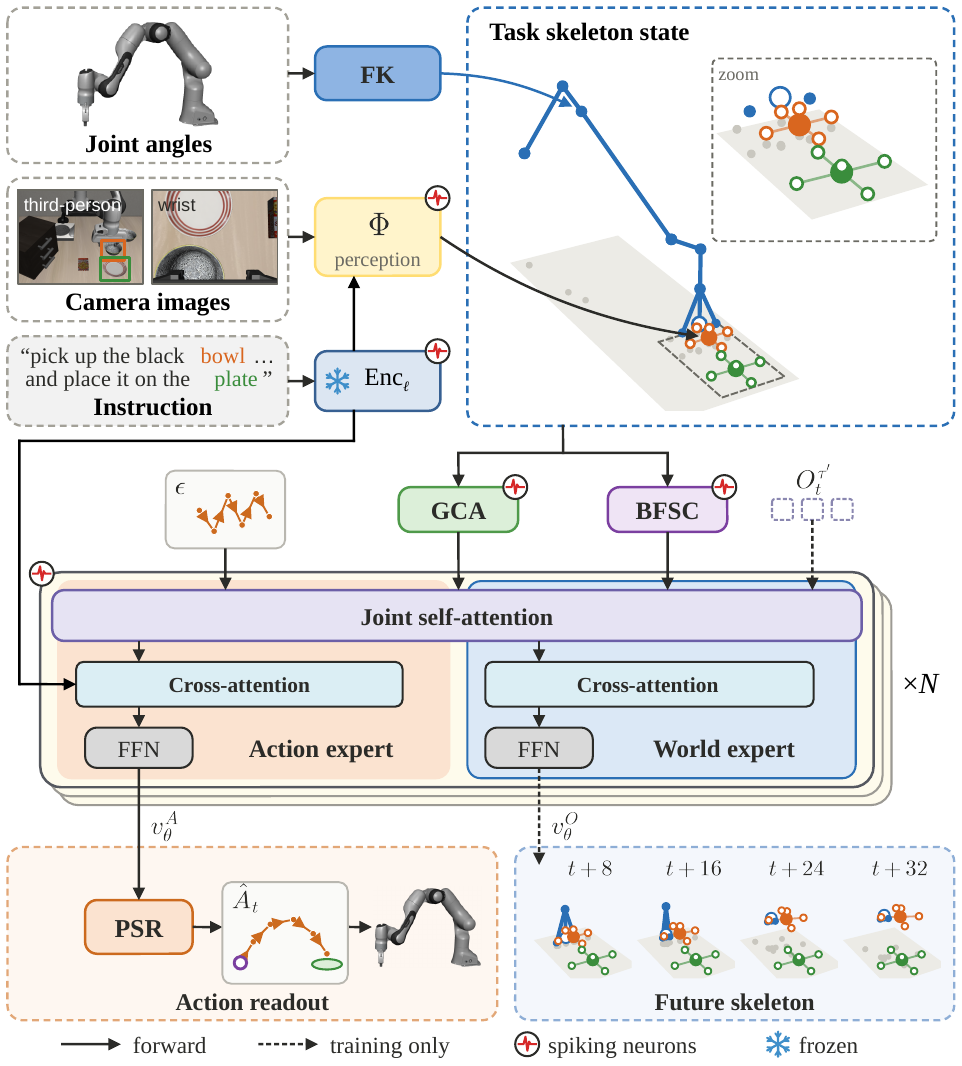}
  \caption{\textbf{Architecture of SpikingWAM.}
  Arm nodes come from forward kinematics, and the spiking front end estimates the object nodes from images.
  The future skeleton is used only in training.}
  \label{fig:arch}
\end{figure}

\begin{figure*}[t]
  \centering
  \includegraphics[width=\textwidth]{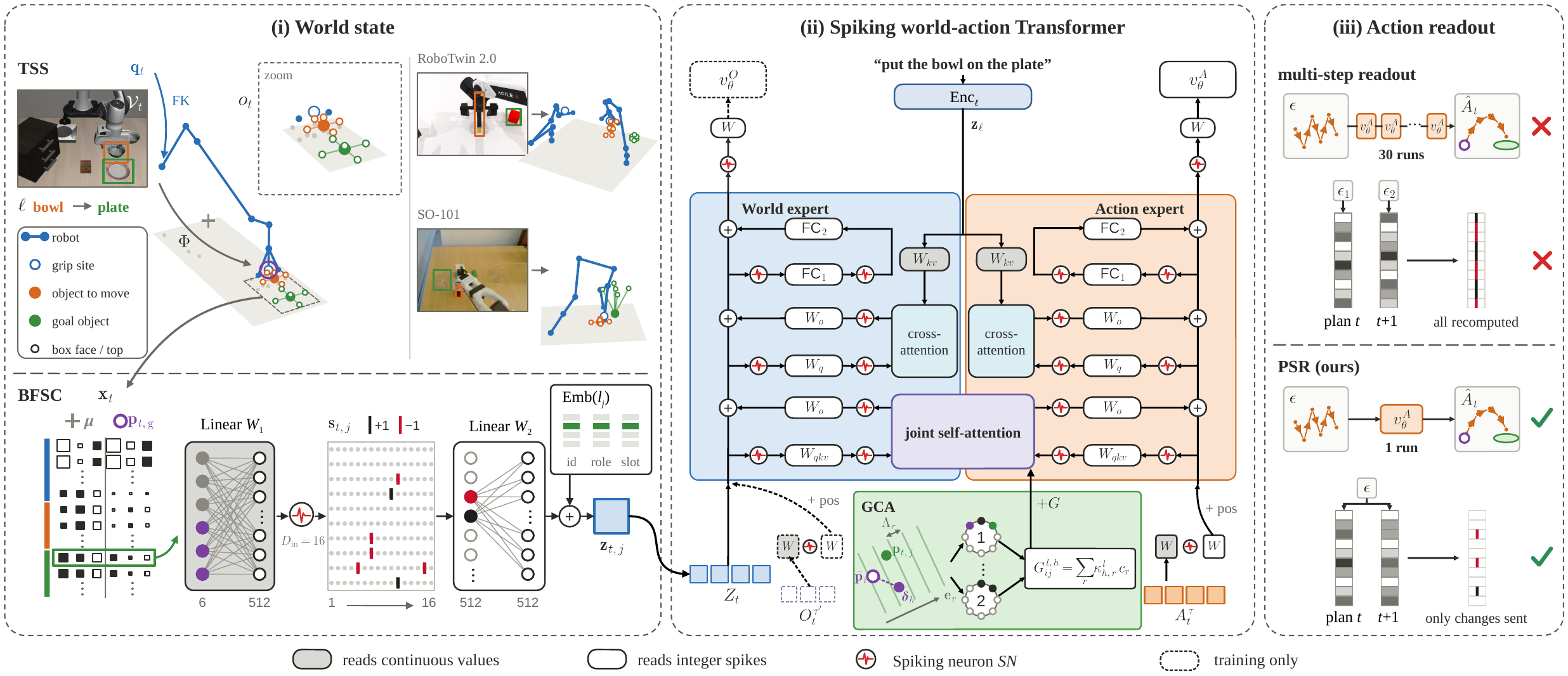}
  \caption{\textbf{SpikingWAM.}
  (i)~The perception front end estimates the keypoints of the task objects from camera images.
  Together with the arm nodes from forward kinematics, they form the current skeleton (TSS), and BFSC writes it into spikes.
  (ii)~The world expert processes the current and future skeletons, and the action expert processes the noisy action chunk.
  The two experts compute self-attention together in every layer, and GCA adds a relative geometry term to it.
  (iii)~At inference the future skeleton is removed.
  The readout of \cref{sec:sfm} runs the network $kN=30$ times per plan and recomputes everything, while PSR runs it once and passes only changes between plans.}
  \label{fig:overview}
\end{figure*}

SpikingWAM addresses the three obstacles to implementing WAMs with SNNs (\cref{app:obstacles}) in three stages along the data flow (\cref{fig:arch,fig:overview}).
\Cref{sec:tss} decides the world state and writes it into spikes, \cref{sec:gca} models the relative geometry of hand and object, and \cref{sec:psr} reads out the action chunk.
\Cref{sec:spikingwam} proposes the paradigm of spiking WAMs and builds SpikingWAM on it from these designs.

\subsection{Task Skeleton State and Body Frame Spike Coding}
\label{sec:tss}
\label{sec:bfsc}

The first stage decides what serves as the world state and writes it into spikes.
When people manipulate objects, their gaze falls almost only on grasp points and targets~\cite{johansson2001eye}.
Inspired by this, TSS keeps only the keypoints of the arm and the task objects,
\begin{equation}
  o_t=\big[\,\mathrm{FK}(\mathbf{q}_t)\,;\ \Phi(\mathcal{V}_t,\ell)\,\big]\in\mathbb{R}^{n\times 3},
  \label{eq:tss}
\end{equation}
where forward kinematics $\mathrm{FK}$ gives the positions of the arm nodes from the joint angles $\mathbf{q}_t$.
The perception front end $\Phi$ is a small spiking network.
It estimates the positions of the task object nodes from the camera images $\mathcal{V}_t$ and the instruction $\ell$ (\cref{app:frontend}).
Besides object centres, TSS includes face centres that rotate with the object, points on moving parts and the two fingertips.
These nodes give the orientation of objects, the state of drawers and doors, and the opening of the gripper.
We write $\mathbf{p}_{t,j}$ for the position of node $j$ and $\iota_j$ for its descriptor, which records its identity, its role and the object it belongs to (\cref{app:model}).
The future world state $O_t$ holds the skeletons of several later frames.
TSS thus lets the policy read only the nodes that matter for the task, and its events concentrate on these nodes.
Future prediction becomes the generation of node coordinates and no longer needs a video generation model.

The skeleton contains only coordinates, so the success of manipulation depends on how precisely the coordinates are written into spikes.
A coordinate component $p$ enters a neuron of \cref{eq:neuron} through the input layer $u=wp+b$.
To avoid saturation over a value range $R$, the layer needs $|w|R/\Vth\le 2D$.
The level spacing of a single neuron therefore satisfies
\begin{equation}
  \Delta p=\frac{\Vth}{|w|}\;\ge\;\frac{R}{2D}.
  \label{eq:step}
\end{equation}
With the workspace as the reference and $D=4$, the level spacing already exceeds 10\,cm (\cref{app:obstacles}).
Inspired by how the cortex represents targets relative to the arm and the hand~\cite{graziano1994coding,buneo2002direct}, BFSC reduces the level spacing in two ways.
Each node also receives its coordinates relative to the grip site, which we call relative coordinates.
The range $R$ of nodes near the hand thus becomes small.
The coordinate input layer uses more levels, $\Din>D$, which raises $D$,
\begin{align}
  \mathbf{x}_{t,j}&=\Big[(\mathbf{p}_{t,j}-\boldsymbol{\mu})\oslash\boldsymbol{\sigma},\ (\mathbf{p}_{t,j}-\mathbf{p}_{t,\mathrm{g}})/\rho\Big],\label{eq:bfsc_in}\\
  \mathbf{z}_{t,j}&=W_{2}\,\SN_{\Din}\big(W_{1}\mathbf{x}_{t,j}\big)+\mathrm{Emb}(\iota_j),\label{eq:bfsc}
\end{align}
where $\mathbf{p}_{t,\mathrm{g}}$ is the grip site, $\boldsymbol{\mu}$ and $\boldsymbol{\sigma}$ are the mean and standard deviation of each axis, $\rho$ is a length scale, and $\mathrm{Emb}(\iota_j)$ embeds the descriptor.
The tokens $\mathbf{z}_{t,j}$ of all nodes form $Z_t$.
The output side also shrinks $R$.
Each step $\mathbf{a}$ of a joint angle action is written as its offset $\mathbf{a}-\mathbf{q}_t$ from the current joint angles (\cref{app:model}).
BFSC thus reduces the level spacing of \cref{eq:step}, so integer spikes keep the millimetre precision that fine manipulation needs.

\subsection{Grid Coincidence Attention}
\label{sec:gca}

The second stage models the world and actions jointly through self-attention.
Actions depend on the relative geometry of hand and object, but \cref{eq:bfsc} quantizes the coordinates of each node separately.
The error of a relative position can therefore reach one level spacing.
GCA is inspired by the periodic firing of grid cells~\cite{hafting2005microstructure} and the coincidence detection of cortical neurons~\cite{konig1996integrator}.
It writes positions as grid spike codes and reads relative geometry directly from the coincidence of codes.
A grid spike code consists of several phase rings, and ring $r$ has a direction $\mathbf{e}_r$, a scale $\Lambda_r$ and $P$ phase neurons,
\begin{equation}
  \psi_{r,\nu}(\mathbf{p})=\mathbb{I}\Big[\big\|\mathbf{e}_r^{\top}\mathbf{p}/\Lambda_r-\nu/P\big\|_{\circ}\le 1/P\Big],
  \label{eq:grid}
\end{equation}
where $\nu=0,\dots,P-1$ and $\|\cdot\|_{\circ}$ is the distance to the nearest integer.
In head $h$ of layer $l$, query $i$ uses $\bar{\mathbf{p}}_i$ as its anchor.
Its geometry term with node $j$ counts the bits that are active in both codes,
\begin{equation}
  G^{l,h}_{ij}=\sum_{r}\kappa^{l}_{h,r}\,\big\langle\boldsymbol{\psi}_r(\bar{\mathbf{p}}_i+\boldsymbol{\delta}_h),\,\boldsymbol{\psi}_r(\mathbf{p}_{t,j})\big\rangle,
  \label{eq:gca}
\end{equation}
where $\boldsymbol{\delta}_h$ is a learnable offset and $\kappa^{l}_{h,r}$ is a learnable gain that starts at 0.
Tokens of the current skeleton use themselves as anchors, and tokens of the future skeleton and the action chunk use the grip site.
The term $G^{l,h}$ is added to the self-attention scores on the nodes of the current skeleton (\cref{app:model}).
The coincidence count is computed by phase directly from the positions given by TSS, so it bypasses the level spacing of \cref{eq:step}.
GCA thus recovers the precision that integer spikes lose in relative geometry, and together with BFSC it keeps millimetre precision.

\subsection{Persistent Single Step Readout}
\label{sec:psr}

The third stage reads the action chunk out of its velocity field $v^{A}_\theta$.
With the readout of several samples in \cref{sec:sfm}, the network runs $kN$ times for each plan.
It draws new noise every time, so consecutive plans cannot reuse computation (\cref{app:obstacles}).
PSR solves the first problem with a one-step readout and the second with a persistent sample stream.

\paragraph{One-step readout.}
The optimal velocity field of \cref{eq:fm} is $v^{*}(A_t^{\tau},\tau\mid c)=\mathbb{E}[\epsilon-A_t\mid A_t^{\tau},c]$.
At $\tau=1$, $A_t^{1}=\epsilon$ is independent of $A_t$, so
\begin{align}
  v^{*}(\epsilon,1\mid c)&=\epsilon-\mathbb{E}[A_t\mid c],\label{eq:vstar}\\
  \hat{A}_t&=\epsilon-v^{A}_{\theta}(\epsilon,1\mid c)\approx\mathbb{E}[A_t\mid c].\label{eq:onestep}
\end{align}
A single integration step from pure noise thus gives the conditional mean of the action chunk.
Training samples noise times that lean toward $\tau=1$~\cite{black2024pi0}, so $v^{A}_\theta$ is trained best at this end.
When the conditional distribution has a single mode, the conditional mean is an executable action.
This condition can be checked offline with samples from the model (\cref{app:readout}).
Otherwise, the same network reads out with several steps as in \cref{sec:sfm}.

\paragraph{Persistent sample stream.}
The conditional mean in \cref{eq:onestep} does not depend on the starting noise.
PSR therefore draws the noise only once per episode, and between two consecutive plans only the skeleton in the network input changes.
The network uses the incremental firing of sigma-delta networks~\cite{oconnor2017sigma} and delta networks~\cite{neil2017delta}.
Each neuron sends only the change of its integer output, and the receiving layer adds it up,
\begin{equation}
  \mathbf{I}^{l}=\mathbf{I}'^{l}+W^{l}\big(\mathbf{s}^{l-1}-\mathbf{s}'^{l-1}\big),
  \label{eq:delta}
\end{equation}
where primed quantities come from the previous run.
\Cref{eq:delta} equals $W^{l}\mathbf{s}^{l-1}$ exactly, but the number of accumulations is only proportional to $\|\mathbf{s}^{l-1}-\mathbf{s}'^{l-1}\|_{1}$ (\cref{app:energy}).
PSR thus runs the network only once per plan and keeps the savings from sparsity.

\subsection{SpikingWAM}
\label{sec:spikingwam}

To address the three obstacles at the start of this section, we propose a sparse and event-driven paradigm for spiking WAMs.
The paradigm builds on the spiking flow matching of \cref{sec:sfm}.
At control time $t$, the instruction $\ell$ and the world state $o_t$ are first encoded separately.
The velocity network $v_\theta$ then gives the velocities of the action chunk and the future world state,
\begin{align}
  \mathbf{z}_\ell&=\mathrm{Enc}_\ell(\ell),\qquad Z_t=\mathrm{Enc}_o(o_t),\label{eq:enc}\\
  v_\theta&=\Big(v^{A}_\theta\big(A_t^{\tau},\tau\mid Z_t,\mathbf{z}_\ell\big),\ v^{O}_\theta\big(O_t^{\tau'},\tau'\mid Z_t,\mathbf{z}_\ell\big)\Big),\label{eq:flow}
\end{align}
where $\mathrm{Enc}_\ell$ is a frozen spiking sentence encoder distilled from MiniLM~\cite{wang2020minilm,reimers2019sentencebert} (\cref{app:model}).
It outputs the sentence vector $\mathbf{z}_\ell$.
The encoder $\mathrm{Enc}_o$ writes the world state into spike tokens $Z_t$.
Each of the two velocities has its own loss in the form of \cref{eq:fm}, and the future term has weight $\lambda_s$.
The network has a world expert that processes the world state and an action expert that processes actions~\cite{fastwam}.
The two experts compute self-attention together in every layer, and the scores on $Z_t$ can include a geometry term $G^{l,h}$.
The tokens of the future world state and the action tokens are hidden from each other, and the two share only the tokens of the current world state.
Each expert then computes cross-attention to the sentence vector and a feedforward network.
The residual stream carries membrane potentials.
Each sublayer fires before its weights~\cite{yao2023spikedriven,yao2024spike} and is modulated by the noise time through adaLN~\cite{peebles2023dit} (\cref{app:model}).
Except for adaLN and the grey modules in \cref{fig:overview}, every weight layer reads only integer spikes.
These layers compute only when a spike fires, and they replace multiply and accumulate with accumulate.
Shiftmax~\cite{wang2025bipolar} normalizes the attention weights with bit shifts (\cref{app:model}).
At inference, the future world state is removed.
A readout obtains the action chunk from $v^{A}_\theta$, and the robot executes its first few steps before it replans.
Each choice of world state $o_t$, encoder $\mathrm{Enc}_o$, geometry term $G^{l,h}$ and readout gives a spiking WAM, where the encoder and the geometry term set the geometric precision.

SpikingWAM uses TSS as the world state, BFSC as the encoder, GCA as the geometry term and PSR as the readout,
\begin{align}
  o_t&=\mathrm{TSS}(\mathbf{q}_t,\mathcal{V}_t,\ell),\qquad Z_t=\mathrm{BFSC}(o_t),\label{eq:spikingwam1}\\
  G^{l,h}&=\mathrm{GCA}(o_t),\qquad \hat{A}_t=\mathrm{PSR}\big(v^{A}_\theta\mid Z_t,\mathbf{z}_\ell\big),\label{eq:spikingwam2}
\end{align}
where the operators follow \cref{eq:tss}, \cref{eq:bfsc_in,eq:bfsc}, \cref{eq:grid,eq:gca} and \cref{eq:onestep,eq:delta}, respectively.
Through the world expert, the loss on the future skeleton shapes the current skeleton features read by the action expert.
Future prediction runs only in training and adds no inference energy.
\section{Experiments}
\label{sec:exp}

\subsection{Experimental Setup}
\label{sec:setup}

\paragraph{Benchmarks and energy.}
LIBERO~\cite{liu2023libero} tests fine manipulation with 40 tasks in four suites, and each task is evaluated for 50 episodes.
LIBERO-Plus~\cite{fei2026liberoplus} applies seven kinds of perturbation to LIBERO in 10,030 variants and tests robustness under perturbations.
RoboTwin 2.0~\cite{chen2025robotwin} tests bimanual manipulation with 50 tasks.
Its Easy setting uses clean scenes, and its Hard setting adds scene randomization.
The evaluation protocols are in \cref{app:protocol}.
We estimate the mean energy per plan for a 45\,nm process~\cite{horowitz20141}, and the perception front end is counted as well.
Each accumulate in a layer that reads spikes costs $\Esop=0.9$\,pJ, and each multiply and accumulate in a layer that reads continuous values costs $\Emac=4.6$\,pJ (\cref{app:energy}).
For other methods, we estimate one forward pass that generates one action chunk, image encoding included.
Their FLOPs come from Spike-driven VLA~\cite{wang2026spikedriven} or from their public architectures.

\paragraph{Model.}
The perception front end and the policy are trained separately on keypoints from the simulator state, and are connected directly at evaluation (\cref{app:frontend}).
The training images of the front end are randomized procedurally and use no LIBERO-Plus assets.
LIBERO and LIBERO-Plus share one model trained on LIBERO demonstrations, which reads out with PSR.
Like the compared methods, the RoboTwin 2.0 model uses 50 clean and about 500 randomized demonstrations per task.
It reads out with the medoid chunk of $N=10$ steps and $k=3$ samples (\cref{sec:analysis}).
The architecture and training are in \cref{app:model}.

\paragraph{Compared methods.}
We divide the compared methods into three groups.
The first group contains ANN VLAs and WAMs with more than 1B parameters, such as $\pi_0$~\cite{black2024pi0}, $\pi_{0.5}$~\cite{intelligence2025pi05}, Fast-WAM~\cite{fastwam} and Cosmos-Policy~\cite{kim2026cosmos}.
The second group contains lightweight ANN VLAs with fewer than 1B parameters, such as TurboVLA~\cite{xie2026turbovla}.
The third group is the spiking VLA Spike-driven VLA~\cite{wang2026spikedriven}.
Their results and parameter counts are taken from the cited papers.
The main results of SkeleWAM use keypoints from the simulator state as input.
The comparison with SpikingWAM in this setting is in \cref{app:skelewam}, and the energy of SkeleWAM is computed on our reproduction (\cref{app:energy}).

\begin{table*}[t]
  \centering
  \caption{\textbf{Success rates (\%) on LIBERO and LIBERO-Plus.}
  Long is LIBERO-10, and Robot is the robot initial state.
  Parameters of SpikingWAM include the text encoder and the perception front end.}
  \label{tab:main}
  \footnotesize
  \setlength{\tabcolsep}{2.6pt}
  \begin{tabular}{@{}llcccccccccccccc@{}}
    \toprule
    & & & \multicolumn{5}{c}{LIBERO} & \multicolumn{8}{c}{LIBERO-Plus} \\
    \cmidrule(lr){4-8}\cmidrule(l){9-16}
    Method & Venue & Params & Spatial & Object & Goal & Long & Avg. & Camera & Robot & Language & Light & Background & Noise & Layout & Total \\
    \midrule
    \multicolumn{16}{@{}l}{\textit{ANN, params $\ge$ 1B}} \\
    OpenVLA~\cite{kim2024openvla} & CoRL'24 & 7.5B & 84.7 & 88.4 & 79.2 & 53.7 & 76.5 & 0.8 & 3.5 & 23.0 & 8.1 & 34.8 & 15.2 & 28.5 & 15.6 \\
    $\pi_0$~\cite{black2024pi0} & RSS'25 & 3.2B & 96.8 & 98.8 & 95.8 & 85.2 & 94.2 & 13.8 & 6.0 & 58.8 & 85.0 & 81.4 & 79.0 & 68.9 & 53.6 \\
    $\pi_{0.5}$~\cite{intelligence2025pi05} & CoRL'25 & 3.4B & 98.8 & 98.2 & 98.0 & 92.4 & 96.9 & 70.6 & 50.5 & 84.4 & 95.7 & 93.4 & 87.4 & 84.1 & 79.7 \\
    OpenVLA-OFT~\cite{kim2025openvlaoft} & RSS'25 & 7.7B & 97.6 & 98.4 & 97.9 & 94.5 & 97.1 & 56.4 & 31.9 & 79.5 & 88.7 & 93.3 & 75.8 & 74.2 & 69.6 \\
    Fast-WAM~\cite{fastwam} & arXiv'26 & 6.7B & 98.2 & \textbf{100.0} & 97.0 & 95.2 & 97.6 & 16.4 & 44.5 & 68.9 & 78.2 & 53.7 & 37.7 & 60.7 & 51.5 \\
    Cosmos-Policy~\cite{kim2026cosmos} & ICLR'26 & 2B & 98.1 & \textbf{100.0} & 98.2 & \textbf{97.6} & \textbf{98.5} & 75.8 & 63.3 & 81.7 & 96.5 & 88.9 & 92.7 & 82.2 & 82.2 \\
    ACoT-VLA~\cite{zhong2026acot} & CVPR'26 & 3.8B & 98.6 & 99.0 & \textbf{99.4} & 97.0 & \textbf{98.5} & 72.6 & 82.6 & \textbf{87.5} & \textbf{97.7} & \textbf{96.5} & 87.8 & \textbf{88.1} & 86.6 \\
    GE-Act~\cite{geact} & ICLR'26 & 2.5B & -- & -- & -- & -- & -- & 60.7 & 77.0 & 77.4 & 95.8 & 86.0 & 90.9 & 80.2 & 80.3 \\
    \midrule
    \multicolumn{16}{@{}l}{\textit{ANN, params $<$ 1B}} \\
    Evo-1~\cite{lin2026evo1} & CVPR'26 & 0.77B & 92.7 & 97.7 & 96.3 & 92.3 & 94.8 & -- & -- & -- & -- & -- & -- & -- & -- \\
    TurboVLA~\cite{xie2026turbovla} & arXiv'26 & 0.2B & 97.0 & \textbf{100.0} & 98.2 & 95.2 & 97.6 & -- & -- & -- & -- & -- & -- & -- & -- \\
    X-VLA~\cite{zheng2025xvla} & ICLR'26 & 0.9B & 98.2 & 98.6 & 97.8 & \textbf{97.6} & 98.1 & -- & -- & -- & -- & -- & -- & -- & -- \\
    \midrule
    \multicolumn{16}{@{}l}{\textit{SNN}} \\
    Spike-driven VLA~\cite{wang2026spikedriven} & arXiv'26 & 0.15B & 95.0 & 95.4 & 96.6 & 82.4 & 92.4 & 41.2 & 46.7 & 48.7 & 84.2 & 61.3 & 34.5 & 67.5 & 53.2 \\
    \rowcolor{ourrow}
    \shadeL\textbf{SpikingWAM} & -- & 0.077B & \textbf{99.0} & \textbf{100.0} & 98.2 & 95.8 & 98.25 & \textbf{94.6} & \textbf{86.5} & 76.1 & 95.2 & 92.9 & \textbf{94.3} & 84.1 & \textbf{88.76}\shadeR \\
    \bottomrule
  \end{tabular}
\end{table*}

\subsection{Main Results}
\label{sec:main}

As shown in \cref{tab:main}, SpikingWAM reaches a mean success of 98.25\% on LIBERO, close to the compared SOTA methods Cosmos-Policy~\cite{kim2026cosmos} and ACoT-VLA~\cite{zhong2026acot}.
It is 5.85 points higher than Spike-driven VLA~\cite{wang2026spikedriven}, which is also a spiking network.
It is also 1.35 and 0.65 points higher than $\pi_{0.5}$~\cite{intelligence2025pi05} and Fast-WAM~\cite{fastwam}, with only 1/44 and 1/86 of their parameters.
SpikingWAM needs only 1.467\,mJ per plan, perception front end included (\cref{fig:teaser}).
Under the same rules, SkeleWAM, an ANN world action model counted from the skeleton interface, needs 92.67\,mJ.
Compared with the policy of SpikingWAM, spiking computation and PSR each save about an order of magnitude (\cref{app:energy}).
Spike-driven VLA, which reads images, needs 20.9\,mJ.

\begin{figure}[t]
  \centering
  \includegraphics[width=\linewidth]{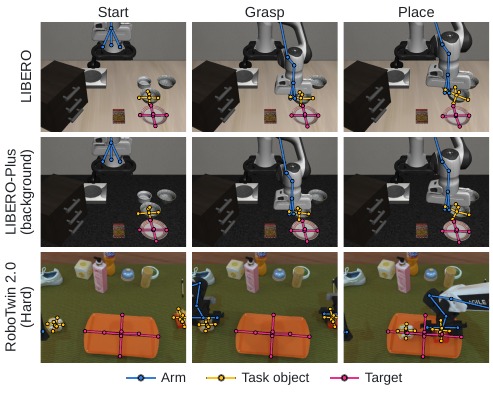}
  \caption{\textbf{Closed-loop runs of SpikingWAM.}
  Each row is one successful episode at the start, the grasp and the last plan before the release, with the task skeleton that the policy reads.
  Object nodes come from the perception front end, and objects that the instruction does not name have no nodes.
  The LIBERO-Plus row changes the background of the same task.}
  \label{fig:rollouts}
\end{figure}

As shown in \cref{tab:main}, SpikingWAM reaches a total success of 88.76\% on LIBERO-Plus, above all compared ANN and SNN methods.
Perturbations of camera, lighting, background and noise change the image, and SpikingWAM estimates only the keypoints of the task objects from the image (\cref{fig:rollouts}).
Its success on these four kinds stays between 92.9\% and 95.2\%, which shows that TSS makes the model focus on the core elements of an embodied scene.

\begin{table}[t]
  \centering
  \caption{\textbf{Success rates (\%) on RoboTwin 2.0.}
  Parameters of SpikingWAM include its text encoder and front end.}
  \label{tab:robotwin}
  \footnotesize
  \setlength{\tabcolsep}{5pt}
  \begin{tabular}{@{}llcccc@{}}
    \toprule
    Method & Venue & Params & Easy & Hard & Avg. \\
    \midrule
    \multicolumn{6}{@{}l}{\textit{Params $\ge$ 1B}} \\
    $\pi_0$~\cite{black2024pi0} & RSS'25 & 3.2B & 65.92 & 58.40 & 62.16 \\
    $\pi_{0.5}$~\cite{intelligence2025pi05} & CoRL'25 & 3.4B & 82.74 & 76.76 & 79.75 \\
    ABot-M0~\cite{yang2026abotm0} & arXiv'26 & 5.9B & 86.06 & 85.08 & 85.57 \\
    GigaWorld-Policy~\cite{ye2026gigaworld} & arXiv'26 & 12.4B & 86.36 & 85.04 & 85.70 \\
    Qwen-VLA~\cite{wang2026qwenvla} & arXiv'26 & -- & 86.10 & \textbf{87.20} & 86.65 \\
    LingBot-VLA~\cite{wu2026lingbotvla} & arXiv'26 & 4.2B & 88.56 & 86.68 & 87.62 \\
    Motus~\cite{bi2025motus} & CVPR'26 & 8.7B & \textbf{88.66} & 87.02 & \textbf{87.84} \\
    \midrule
    \multicolumn{6}{@{}l}{\textit{Params $<$ 1B}} \\
    X-VLA~\cite{zheng2025xvla} & ICLR'26 & 0.9B & 72.80 & 72.84 & 72.82 \\
    \rowcolor{ourrow}
    \shadeL\textbf{SpikingWAM} & -- & 0.104B & 83.02 & 77.38 & 80.20\shadeR \\
    \bottomrule
  \end{tabular}
\end{table}

We evaluate SpikingWAM on the 50 bimanual tasks of RoboTwin 2.0.
Bimanual manipulation in more complex scenes needs more accurate skeletons, so we enlarge the perception front end to 30M (\cref{app:frontend}).
As shown in \cref{tab:robotwin}, SpikingWAM reaches a mean success of 80.20\% with 0.104B parameters.
It is 0.45 and 7.38 points higher than $\pi_{0.5}$~\cite{intelligence2025pi05} and X-VLA~\cite{zheng2025xvla}, with only 1/33 and 1/9 of their parameters.
All compared methods use extra pretraining on robot data, while the SpikingWAM policy is trained from scratch only on RoboTwin 2.0 demonstrations.
This shows that SpikingWAM, with the task skeleton as its world state, extends to bimanual manipulation and keeps competitive robustness under scene randomization.

\subsection{Analysis}
\label{sec:analysis}

This section examines the problems that TSS and BFSC target, and when the one-step readout applies.

\begin{figure}[t]
  \centering
  \includegraphics[width=\linewidth]{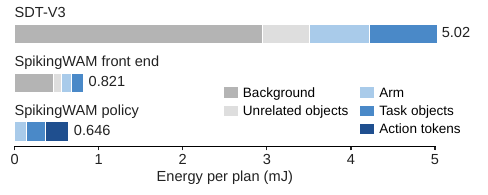}
  \caption{\textbf{Energy per plan by region on LIBERO.}
  The energy of each model is split in proportion to its events in each region (\cref{app:events}).
  SDT-V3 encodes the two camera images.}
  \label{fig:energy_regions}
\end{figure}

\paragraph{Task skeleton.}
With images as the world state, most events are spent on a background unrelated to the task, and small jitter in the image keeps triggering new events.
As shown in \cref{fig:energy_regions}, 58.7\% of the events of the spiking vision backbone SDT-V3~\cite{yao2025scaling} fall on the background.
The policy of SpikingWAM has no events on the background or on unrelated objects.
The perception front end only needs to estimate a few keypoints from the images.
Its energy per plan is about 1/6 of that of SDT-V3.
With one grey level of noise per camera frame, the energy of SDT-V3 rises by 0.64\,mJ, while that of the front end rises by only 0.076\,mJ.
With the front end included, SpikingWAM spends 0.46\,mJ per plan on the background, while SDT-V3 spends 2.95\,mJ.
In the policy, 75\% of the attention of action tokens falls on the task objects, above the 57\% share of task object nodes (\cref{app:events}).
This shows that TSS focuses the policy's events and attention on the parts that matter for the task, which brings out the low power advantage of event-driven SNNs.

\paragraph{Coordinate precision.}
Encoding coordinates with integer spikes loses precision, and the loss happens in the input layer where coordinates are written into spikes (\cref{app:obstacles}).
We read the coordinates back linearly from the spikes of this layer.
With BFSC, the error of absolute coordinates comes close to that of an ANN model with the same architecture and continuous input, and the error of relative coordinates is only 3.7 to 4.8\,mm.
Without BFSC, the error of absolute coordinates grows to 17 to 24\,mm.
This shows that BFSC recovers the loss from encoding coordinates with integer spikes, so the SNN keeps the millimetre precision that fine manipulation needs.

\paragraph{When the one-step readout applies.}
The one-step readout gives the conditional mean of the action chunk.
This mean is an executable action only when it lies among the model's own samples.
We check this condition offline with a ratio.
It divides the distance from the mean to its nearest sample by the nearest neighbour distance among the samples (\cref{app:readout}).
The ratio is at most 1.00 on every LIBERO task and at least 2.00 on every RoboTwin 2.0 task, so RoboTwin 2.0 reads out with the medoid chunk of 10 steps.
Closed loop results agree with the check.
On LIBERO, the one-step readout succeeds more often than the medoid chunk with 10 steps (\cref{sec:ablation}).
This shows that the offline check tells whether a benchmark suits the one-step readout without closed loop rollouts.
When it does, each plan runs the network only once, which saves time and energy.

\subsection{Real Robot Experiments}
\label{sec:real}

\begin{figure}[t]
  \centering
  \includegraphics[width=\linewidth]{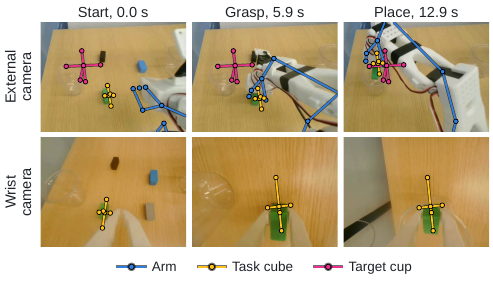}
  \caption{\textbf{One successful run of SpikingWAM on the SO-101 arm.}
  The instruction is ``Put the green cube into the left cup.''
  The columns show the start, the grasp and the moment before the cube enters the cup.
  The top row shows the task skeleton on the external camera, and the bottom row shows the cube nodes that the front end estimates from the wrist camera.
  The other cubes and the right cup have no nodes.}
  \label{fig:real}
\end{figure}

To test whether SpikingWAM also works in real scenes, we run a pick and place task on an SO-101 arm.
The instruction is ``Put the green cube into the left cup.''
The object keypoints are estimated from camera images by spiking networks.
SpikingWAM, X-VLA~\cite{zheng2025xvla} and $\pi_{0.5}$~\cite{intelligence2025pi05} are trained on the same demonstrations, and each model runs 10 trials (\cref{app:real}).
The success rate of SpikingWAM lies between those of the two models, and its energy, inference time and memory are far lower (\cref{tab:real}).
SpikingWAM thus performs pick and place efficiently on a real robot (\cref{fig:real}).

\begin{table}[t]
  \centering
  \caption{\textbf{Successes and cost on the real robot.}
  Each model runs 10 trials.
  Energy and time are per plan, and time and peak GPU memory are measured in BF16 on the same H100 (\cref{app:real}).}
  \label{tab:real}
  \footnotesize
  \setlength{\tabcolsep}{3.5pt}
  \begin{tabular}{@{}lcccc@{}}
    \toprule
    Model & Successes & Energy (mJ) & Time (ms) & Memory (GiB) \\
    \midrule
    X-VLA~\cite{zheng2025xvla} & 5 / 10 & 3,660 & 63 & 3.5 \\
    $\pi_{0.5}$~\cite{intelligence2025pi05} & \textbf{8 / 10} & 8,740 & 137 & 7.1 \\
    \rowcolor{ourrow}
    \shadeL\textbf{SpikingWAM} & 7 / 10 & \textbf{3.07} & \textbf{4.4} & \textbf{0.63}\shadeR \\
    \bottomrule
  \end{tabular}
\end{table}

\subsection{Ablation Study}
\label{sec:ablation}

\begin{table}[t]
  \centering
  \caption{\textbf{Ablation of BFSC and GCA.}
  Object keypoints come from the perception front end, and the last row is SpikingWAM.
  Two objects covers the 6 LIBERO-10 tasks that place two objects.}
  \label{tab:ablation}
  \footnotesize
  \setlength{\tabcolsep}{2.5pt}
  \begin{tabular}{@{}cccccc@{}}
    \toprule
    \multicolumn{2}{c}{BFSC} & & \multicolumn{3}{c}{Success (\%)} \\
    \cmidrule(r){1-2}\cmidrule(l){4-6}
    Relative coords. & Input layer & GCA & LIBERO & LIBERO-10 & Two objects \\
    \midrule
    \xmark & \xmark & \xmark & 91.45 & 77.4 & 70.7 \\
    \xmark & \xmark & \cmark & 92.00 & 77.4 & 70.3 \\
    \cmark & \xmark & \xmark & 95.60 & 88.2 & 82.0 \\
    \cmark & \cmark & \xmark & 97.80 & 93.6 & 93.3 \\
    \rowcolor{ourrow}
    \shadeL\cmark & \cmark & \cmark & \textbf{98.25} & \textbf{95.8} & \textbf{95.0}\shadeR \\
    \bottomrule
  \end{tabular}
\end{table}

\begin{table}[t]
  \centering
  \caption{\textbf{Readout and energy.}
  One LIBERO model with object keypoints from the perception front end.
  $N$ is the number of integration steps, and the sample mean averages $k$ action chunks.
  Energy covers only the policy, and the value for $k=36$ is scaled from the measured breakdown (\cref{app:readout}).}
  \label{tab:readout}
  \footnotesize
  \setlength{\tabcolsep}{5pt}
  \begin{tabular}{@{}lccccc@{}}
    \toprule
    & & & & \multicolumn{2}{c}{Success (\%)} \\
    \cmidrule(l){5-6}
    Readout & $N$ & $k$ & Energy (mJ) & LIBERO & LIBERO-Plus \\
    \midrule
    Medoid & 10 & 3 & 7.48 & 91.20 & 82.93 \\
    Sample mean & 10 & 3 & 7.48 & 96.15 & 87.02 \\
    Sample mean & 10 & 36 & 85.85 & 98.10 & -- \\
    Single sample & 1 & 1 & 1.04 & 97.70 & -- \\
    \rowcolor{ourrow}
    \shadeL\textbf{PSR} & 1 & 1 & \textbf{0.646} & \textbf{98.25} & \textbf{88.76}\shadeR \\
    \bottomrule
  \end{tabular}
\end{table}

This section ablates BFSC, GCA, future prediction and the readout.
As shown in \cref{tab:ablation}, relative coordinates and the coordinate input layer raise the LIBERO success rate from 91.45\% to 97.80\%, and the success on the two object tasks from 70.7\% to 93.3\%.
With both in place, GCA raises LIBERO-10 from 93.6\% to 95.8\% at only 2.18\,$\mu$J per plan.
Future prediction raises LIBERO-Plus by 1.55 points, mainly under language and robot initial state perturbations (\cref{app:future}).
As shown in \cref{tab:readout}, the sample mean succeeds more often than the medoid chunk at the same energy, so the conditional mean suits execution better than a single sample.
The one-step readout gives the conditional mean directly, so a single sample matches the mean of 36 samples with 10 steps.
The persistent sample stream lowers the energy further, and PSR has the highest success rate among all readouts.
This shows that all these components are effective.
Each raises the success rate or lowers the energy per plan, and together they let an SNN perform fine manipulation at low power.
\section{Conclusion}
\label{sec:conclusion}

We proposed a sparse and event-driven paradigm for spiking world action models and built on it SpikingWAM, the first spiking world action model.
TSS uses only the keypoints of the arm and the task objects as the world state, and BFSC and GCA keep its millimetre precision in integer spikes.
PSR runs the network once per plan and passes only the changes between plans.
On simulated benchmarks, SpikingWAM achieves competitive success rates with far less energy than ANN methods, and it also works on a real robot arm.
Our results show that the choice of world state is key to spiking WAMs.
A task skeleton lets the network spend its events on the arm and the task objects, so its sparsity translates into low energy.
We hope this paradigm helps bring WAMs to neuromorphic hardware and advances low power embodied intelligence.
{
    \small
    \bibliographystyle{ieeenat_fullname}
    \bibliography{refs}
}

\maketitlesupplementary
\appendix
\counterwithin{equation}{section}
\counterwithin{table}{section}
\counterwithin{figure}{section}

\section{Analysis of the Three Obstacles}
\label{app:obstacles}

The designs of \cref{sec:method} target three parts of the model, namely the world state, geometric precision and the readout.
This appendix derives the cost of each part from \cref{eq:neuron} and gives measured diagnostics.

\paragraph{Composition of energy.}
By \cref{eq:unfold}, each unit spike fired in layer $l$ makes the weight layers that read it perform $m_l$ accumulations.
Here $m_l$ is the sum of the output dimensions of these layers.
Without reuse across consecutive runs, the energy of one plan is
\begin{equation}
  E=kN\Big(\Esop\sum_{j=1}^{n}\sum_{l}m_{l}\big\|\mathbf{s}^{l}_{j}\big\|_{1}+\Emac\,M\Big),
  \label{eq:energy}
\end{equation}
where $n$ is the number of tokens and $\mathbf{s}^{l}_{j}$ holds the spikes fired by token $j$ in layer $l$.
$M$ is the number of MACs in the layers that read continuous values during one run, and $\Esop$ and $\Emac$ are the energies of one accumulate and one MAC (\cref{sec:setup}).
In \cref{eq:energy}, the sum over tokens depends on the world state, and the outer factor $kN$ depends on the readout.

\paragraph{World state.}
The sum over tokens in \cref{eq:energy} depends on how the world state is represented.
With images as the world state, $n$ is the number of image patches.
Events spread over the image area, and most of the image is background.
Small jitter in the image also triggers new events on the patches.
For a skeleton, $n$ is only the number of nodes of the arm and the task objects.
The events of the policy therefore fall only on these nodes and on the action tokens.
\Cref{fig:energy_regions} and \cref{app:events} give the measurements for both world states.

\paragraph{Geometric precision.}
The skeleton compresses the scene into coordinates, so the success of manipulation depends on their precision.
A coordinate component $p$ enters a neuron of \cref{eq:neuron} through the input layer $u=wp+b$.
To avoid saturation over a range $R$, the neuron needs $|w|R/\Vth\le 2D$.
The coordinate difference between two adjacent levels therefore satisfies
\begin{equation}
  \Delta p=\frac{\Vth}{|w|}\;\ge\;\frac{R}{2D}.
  \label{eq:step_app}
\end{equation}
The three axes of the LIBERO workspace span 1.0 to 1.4\,m.
With $D=4$, the level spacing of a single neuron is 12.5 to 17.5\,cm.
The input layer encodes each coordinate with many neurons.
Let S0 be a spiking skeleton model without our designs, with absolute coordinates as input and $D=4$.
Reading coordinates back from the spikes of its input layer still gives an error of 17 to 24\,mm.
The error is 7.5 to 10.5\,mm for the ANN model R0 with continuous input and 9.4 to 9.9\,mm for SpikingWAM (\cref{sec:analysis}).
S1 adds relative coordinates to S0.
On S1, removing the rounding of all inner layers at inference leaves the action error unchanged, at 0.0362 to 0.0378 against 0.0366 with rounding.
SpikingWAM gives the same result, so the precision loss happens in the input encoding layer.
This loss falls on tasks that need precise placement.
S0 reaches 70.7\% on the 6 tasks that must place two objects and 95.1\% on the other 34 tasks (\cref{tab:ablation}).

By \cref{eq:step_app}, both a smaller $R$ and a larger $D$ shrink the level spacing, and BFSC does both (\cref{sec:tss}).
The relative position of hand and object is the difference of two separately quantized coordinates, so its error can reach one level spacing.
GCA obtains relative geometry directly from the coincidence of their spike codes.

\paragraph{Readout.}
In \cref{eq:energy}, the factor $kN$ multiplies the whole energy.
With $N=10$ integration steps and the medoid of $k=3$ samples, each plan runs the network 30 times, and 95\% of the energy is in the action branch (\cref{app:energy}).
Incremental firing can reuse what stays the same between consecutive integration steps of one sample.
However, each plan draws new noise, so the whole input of the action branch changes, and nothing can be reused between consecutive plans.
PSR reduces the readout to one step and one sample and then fixes the starting noise, so only changes pass between consecutive plans.
\Cref{tab:readout} gives the energy of each readout.

\section{Perception Front End}
\label{app:frontend}

The perception front end $\Phi$ of LIBERO (\cref{eq:tss}) estimates the positions of the task object nodes from the third-person and wrist RGB images.
The arm nodes still come from forward kinematics.

\paragraph{Architecture.}
Both images are resized to $160\times160$ and encoded by one shared spiking convolutional backbone.
Convolutions and depthwise separable residual blocks bring the resolution down to 1/16 step by step.
Two spiking self-attention layers~\cite{zhou2023spikformer} run on top of it.
Their output is upsampled and merged with the 1/8 features, which gives a $20\times20$ feature map for each view.
All layers use $D=4$ and fire only non-negative spikes.
The first convolution reads pixels, and all other layers read only spikes.
The front end has 3.46M parameters.

\paragraph{Queries and readout.}
Each object node has a query vector.
It is the sum of the embeddings of the descriptor $\iota_j$ and of the object class, plus a linear map of the sentence vector $\mathbf{z}_\ell$, the gripper position and opening, and the camera matrix of the view.
The object class comes from the task definition, and the camera matrix holds the known intrinsics and extrinsics.
The query takes an inner product with every cell of the feature map.
Two spiking fully connected layers then read out three distributions over one axis each.
They cover the horizontal pixel coordinate, the vertical pixel coordinate and the height of the node, which is coordinate classification~\cite{li2022simcc}.
We take the expectation of each distribution and project the pixel along its camera ray onto the horizontal plane at this height.
This gives the position in the robot frame.
The positions from the two views are fused with weights given by the softmax of their confidences.

\paragraph{Training.}
The training images come from rendering all LIBERO demonstrations again with the renderer of the closed loop evaluation.
The node positions in the simulator state serve as labels.
Besides the original scenes, three more groups of images replace the textures of the table, floor and walls, change the lights and turn the third-person camera.
They also move the objects on the table that are not near the gripper at random, which follows domain randomization~\cite{tobin2017domain}.
All textures are generated procedurally, and no LIBERO-Plus assets are used.
During training, half of the third-person images get blur, fog or sensor noise, and instructions are rewritten from a word list.
The last 10\% of the demonstrations of each task are held out for validation, and the training set has 627k frames.
The loss sums three terms.
They are the cross entropy of the three distributions against Gaussian soft labels, the L1 errors of the positions from each view and after fusion, and the binary cross entropy of the confidence.
We train with AdamW, learning rate $10^{-3}$ with cosine decay and batch size 128 for 200k steps.
The front end is trained alone, and the policy is not fine-tuned on estimated skeletons.

\paragraph{Errors and closed loop results.}
On the held out demonstrations in the original scenes, the median error of object nodes is 0.28\,cm and the 90th percentile is 1.20\,cm.
Of these nodes, 87.8\% lie within 1\,cm.
The two rows of \cref{tab:frontend} use the same policy and change only the source of the object keypoints.
They differ by 0.10 points on LIBERO.
On LIBERO-Plus they differ by 1.65 points, and the gap is largest on background, layout and robot initial state.

\begin{table}[h]
  \centering
  \caption{\textbf{Closed loop success (\%) with object keypoints from the simulator state (\S) and from the perception front end.}
  The policy is the same in both columns.}
  \label{tab:frontend}
  \footnotesize
  \setlength{\tabcolsep}{6pt}
  \begin{tabular}{@{}lc>{\columncolor{ourrow}[\tabcolsep][0pt]}c@{}}
    \toprule
    & Simulator state (\S) & \multicolumn{1}{c@{}}{\textbf{Perception front end}} \\
    \midrule
    \emph{LIBERO} & & \\
    \quad Spatial & 98.4 & 99.0 \\
    \quad Object & 99.8 & 100.0 \\
    \quad Goal & 99.0 & 98.2 \\
    \quad LIBERO-10 & 96.2 & 95.8 \\
    \quad Average & 98.35 & 98.25 \\
    \midrule
    \emph{LIBERO-Plus} & & \\
    \quad Camera & 96.1 & 94.6 \\
    \quad Robot initial state & 88.5 & 86.5 \\
    \quad Language & 77.6 & 76.1 \\
    \quad Light & 95.3 & 95.2 \\
    \quad Background & 95.8 & 92.9 \\
    \quad Noise & 95.3 & 94.3 \\
    \quad Layout & 86.8 & 84.1 \\
    \quad Total & 90.41 & 88.76 \\
    \bottomrule
  \end{tabular}
\end{table}

\paragraph{RoboTwin 2.0.}
The RoboTwin 2.0 front end reads the head camera and the two wrist cameras.
It has the same structure as the front end above, with wider layers.
Its training images come from expert demonstrations and from closed loop episodes with the front end, with the node positions in the simulator state as labels.
\Cref{tab:frontend_rt} compares front ends of different sizes with the same policy.
From 7.4M to 30.1M parameters, the mean success rate rises by 13.2 points, and the energy per plan rises by 7.9\,mJ.
At 30.1M, the front end and the policy use about the same energy, 20.6\,mJ in total.
This size gives the highest success rate, so the main text uses it.
The front end of the main text has this size and is trained further on images of closed-loop episodes with more backgrounds.
With 100 episodes per task, it reaches 83.02\% on Easy and 77.38\% on Hard (\cref{tab:robotwin}), with 20.7\,mJ per plan.

\begin{table}[h]
  \centering
  \caption{\textbf{Perception front ends of different sizes on RoboTwin 2.0.}
  The policy is the same, and each task runs 10 episodes.
  Energy is the mean per plan and includes the 10.7\,mJ of the policy.
  The shaded row is the size used in the main text.
  \S{} marks object keypoints from the simulator state.}
  \label{tab:frontend_rt}
  \footnotesize
  \setlength{\tabcolsep}{4pt}
  \begin{tabular}{@{}lcccc@{}}
    \toprule
    Front end & Energy (mJ) & Easy & Hard & Avg. \\
    \midrule
    7.4M & 12.7 & 74.2 & 59.0 & 66.6 \\
    12.5M & 14.7 & 80.6 & 69.8 & 75.2 \\
    21.0M & 17.5 & 81.6 & 73.2 & 77.4 \\
    \rowcolor{ourrow}
    \shadeL 30.1M & 20.6 & 83.8 & 75.8 & 79.8\shadeR \\
    \midrule
    Simulator state (\S) & 10.7 & 89.0 & 87.8 & 88.4 \\
    \bottomrule
  \end{tabular}
\end{table}

\section{Model Details}
\label{app:model}

\paragraph{Nodes.}
On LIBERO, the arm has 11 nodes, namely the origins of Panda links 1 to 7 and of the hand, the grip site and the two fingertips.
On RoboTwin 2.0, each of the two arms has 10 nodes, namely the origins of the base link and of links 1 to 6, the grip site and the two fingertips.
Each task object adds a group of nodes.
They are its centre, the centres of its top face and four side faces, which rotate with the object, its functional points, and the centres of moving parts such as drawers, doors and knobs.
RoboTwin 2.0 also has contact points, target points and orientation points.
Coordinates are in the robot frame.
On RoboTwin 2.0 they are in the world frame that holds the bases of both arms.
The descriptor $\iota_j$ is the sum of three embeddings, for the node index, the role and the slot of the object.
The same index means the same part in every episode.
The future skeleton has one token per frame.

\paragraph{Actions.}
On LIBERO, the action is the change of the end effector pose plus the gripper command.
On RoboTwin 2.0, the action holds the joint angles and grippers of both arms.
The joint angles are written as offsets from the current joint angles $\mathbf{q}_t$, and $\mathbf{q}_t$ is added back outside the network.
On RoboTwin 2.0, each link node of the arms has one more channel with the measured angle of its joint.
The network thus sees the posture to which the action chunk is relative.

\paragraph{Hyperparameters.}
Spiking neurons use $D=4$, and the coordinate input layer uses $\Din=16$.
The initial threshold of $\SN_{\Din}$ is scaled by $D/\Din$, so the range stays the same while the number of levels grows.
The length scale of relative coordinates is $\rho=0.2$\,m, and in bimanual tasks each hand gives its own relative coordinates.
The grid spike code has 6 directions and 12 scales, which start at 1\,cm and grow by a factor of 1.5.
Each ring has 8 phase neurons, so the code has 72 rings.
Rings with small scales resolve millimetres, and rings with large scales tell near from far.
The offset $\boldsymbol{\delta}_h$ of each GCA head starts from a normal distribution with a standard deviation of 3\,cm, and the gain $\kappa$ starts at 0.
In bimanual tasks, each hand takes half of the GCA heads.
Shiftmax~\cite{wang2025bipolar} rounds attention scores to multiples of 1/4 and rounds the denominator to a power of 2.

\paragraph{Layer structure.}
The residual stream $\mathbf{h}$ carries membrane potentials, and each sublayer fires before its weights~\cite{yao2023spikedriven,yao2024spike},
\begin{equation}
  \mathbf{h}\leftarrow\mathbf{h}+\boldsymbol{\gamma}\odot\mathcal{F}\big(\SN(\mathbf{u})\big),\qquad \mathbf{u}=(1+\boldsymbol{\beta})\odot\mathrm{LN}(\mathbf{h})+\boldsymbol{\alpha},
  \label{eq:layer}
\end{equation}
where $\mathcal{F}$ is the sublayer (self-attention, cross-attention or feedforward network) and $\boldsymbol{\alpha},\boldsymbol{\beta},\boldsymbol{\gamma}$ are modulation coefficients that adaLN~\cite{peebles2023dit} computes from the noise time.

\paragraph{Architecture and training.}
The two benchmarks use the same architecture.
They differ in their nodes, their action representation and the amount of training (\cref{tab:model}).
The spiking sentence encoder is distilled from MiniLM, stays frozen during policy training and runs once per episode.

\begin{table}[h]
  \centering
  \caption{\textbf{Architecture and training.}}
  \label{tab:model}
  \footnotesize
  \setlength{\tabcolsep}{3pt}
  \begin{tabular}{@{}lp{0.37\linewidth}p{0.37\linewidth}@{}}
    \toprule
    & LIBERO & RoboTwin 2.0 \\
    \midrule
    Experts & \multicolumn{2}{p{0.76\linewidth}}{world expert and action expert, width 512, 6 layers, 8 heads, feedforward width 3.25 times, block mask} \\
    Parameters & policy 51.3M, spiking text encoder 22.7M, perception front end 3.46M & policy 51.5M, spiking text encoder 22.7M, perception front end 30.1M \\
    Horizon & \multicolumn{2}{p{0.76\linewidth}}{action chunk of 32 steps with the first 16 executed; future skeleton of 8 frames at 4, 8, \dots, 32 steps ahead} \\
    Loss & \multicolumn{2}{p{0.76\linewidth}}{independent noise times for the action and the future skeleton, $\lambda_s=1$} \\
    Training & 60k steps, batch 48, AdamW, learning rate $2\times10^{-4}$ with cosine decay to $2\times10^{-5}$, EMA 0.999, bf16 & 150k steps, batch 512, learning rate $6.5\times10^{-4}$ with cosine decay to $2\times10^{-5}$, others as LIBERO \\
    Data & 1,726 demonstrations of the four suites, 279,498 steps & 2,500 clean and 24,600 randomized demonstrations, 6,046,558 steps, 1,023,627 instructions seen in training \\
    \bottomrule
  \end{tabular}
\end{table}

\section{Derivation of the Readout}
\label{app:readout}

\paragraph{One-step readout and the sample mean.}
In \cref{eq:vstar}, $A_t^{1}=\epsilon$ is independent of $A_t$ given $c$.
Conditioning on $A_t^{1}$ therefore fixes only $\epsilon$ and leaves the distribution of $A_t$ unchanged, so $v^{*}(\epsilon,1\mid c)=\epsilon-\mathbb{E}[A_t\mid c]$.
One integration step of size 1 from $\tau=1$ then gives $\mathbb{E}[A_t\mid c]$.
Samples integrated over many steps roughly follow the conditional distribution that the model learns.
The mean of $k$ samples therefore tends to the same conditional mean as $k\to\infty$.
Training draws the noise time from $\mathrm{Beta}(1.5,1)$, which leans toward $\tau=1$.

\paragraph{Offline check of the one-step readout.}
For each task we take 32 training states.
For each state we draw 12 samples of 10 steps with independent noise and one chunk from the one-step readout.
We use the arm dimensions of the first 16 executed steps.
The ratio divides the distance from a chunk to its nearest sample by the median distance from each sample to its nearest neighbour (\cref{tab:check}).
When the samples form one cluster, the sample mean gives a ratio of about 0.7.
On LIBERO, the gripper samples split into two groups by the closing time, and the one-step readout gives a gripper ratio of 7.84.
Execution uses only the sign of the gripper command, so this does not affect it.

\begin{table}[h]
  \centering
  \caption{\textbf{Offline check of the one-step readout.}
  Mean over tasks, with the range in brackets.}
  \label{tab:check}
  \footnotesize
  \setlength{\tabcolsep}{4pt}
  \begin{tabular}{@{}lcc@{}}
    \toprule
    & LIBERO (40 tasks) & RoboTwin 2.0 (50 tasks) \\
    \midrule
    Mean of 12 samples & 0.67 (0.63--0.70) & 1.05 (0.73--2.95) \\
    One-step readout & 0.88 (0.81--1.00) & 3.10 (2.00--5.07) \\
    \bottomrule
  \end{tabular}
\end{table}

\paragraph{Paired episodes of the readouts.}
The readouts in \cref{tab:readout} are compared episode by episode (notation in \cref{app:protocol}).
The sample mean with 10 steps against the medoid chunk ($k=3$) gives 155\,:\,56.
The one-step single sample against the mean of 36 samples with 10 steps gives 30\,:\,38.
PSR against a one-step single sample with new noise in each plan gives 41\,:\,30.
PSR against the medoid chunk gives 170\,:\,29 on LIBERO and 1,001\,:\,416 on LIBERO-Plus.

\section{Energy Counting}
\label{app:energy}

\paragraph{Energy model and counting rules.}
Each accumulate in a layer that reads spikes costs $\Esop=0.9$\,pJ, and each MAC in a layer that reads continuous values costs $\Emac=4.6$\,pJ, for a 45\,nm process~\cite{horowitz20141}.
Weight layers that read only spikes are counted by the accumulates that unit spikes trigger.
Layers that read continuous values, layer normalization and adaLN modulation are counted as MACs.
The policy and the perception front end both run with incremental firing (\cref{eq:delta}), so a layer accumulates only the weight columns of the spikes that changed.
The first convolution of the front end reads pixels and is counted as MACs.
GCA follows the same rule.
Each query spike performs one AND with each key, and each coincidence adds one gain in each layer.
Each of these counts as one accumulate, and together they cost 2.18\,$\mu$J per plan.
Energy is the mean per plan.
On LIBERO it is averaged over 446 plans on the demonstration trajectories of the 40 tasks, with object keypoints from the perception front end.
On RoboTwin 2.0, the policy is averaged over 708 plans on the demonstration trajectories of the 50 tasks, and the perception front end over 2,150 plans of closed-loop episodes.
The instruction is encoded only once per episode.
The spiking MiniLM needs 0.578\,mJ, and the text keys and values of the action expert need another 21.7\,$\mu$J.
This part is not counted in the energy per plan.

\begin{table}[h]
  \centering
  \caption{\textbf{Energy per plan of the final LIBERO policy.}
  Accumulates (SOP) and MACs are counted in millions.}
  \label{tab:energy}
  \footnotesize
  \setlength{\tabcolsep}{3pt}
  \begin{tabular}{@{}lccccc@{}}
    \toprule
    & \multicolumn{2}{c}{Skeleton} & \multicolumn{2}{c}{Action} & \\
    \cmidrule(lr){2-3}\cmidrule(lr){4-5}
    Counting & SOP & MAC & SOP & MAC & Energy (mJ) \\
    \midrule
    Full computation & 781.7 & 2.12 & 738.4 & 2.70 & 1.390 \\
    Skipping unchanged tokens & 731.2 & 2.12 & 687.2 & 2.47 & 1.298 \\
    Incremental firing & 388.0 & 2.04 & 305.3 & 2.39 & 0.644 \\
    \rowcolor{ourrow}
    \shadeL\textbf{Incremental firing + GCA} & \multicolumn{4}{c}{adds 2.42M accumulates} & \textbf{0.646}\shadeR \\
    \bottomrule
  \end{tabular}
\end{table}

Skipping the tokens whose input is unchanged is the counting rule for ANN models, and SkeleWAM is counted with this rule.

\paragraph{Source of the energy gap.}
The original SkeleWAM paper~\cite{sheng2026skelewam} does not release its node definition or its text encoder.
We reproduce SkeleWAM with the architecture of the original paper, which is R0 in \cref{app:obstacles}.
The reproduction uses our node definition and frozen MiniLM token features, and it needs 92.67\,mJ per plan.
On one SpikingWAM model, this gap of about 144 times can be measured step by step.
With the same medoid readout at $k=3$ and $N=10$, spiking computation and incremental firing lower the energy from 92.67\,mJ to 7.48\,mJ (12.4 times).
The one-step readout with a single sample lowers it further to 1.04\,mJ (7.2 times), and the persistent sample stream lowers it to 0.646\,mJ (1.6 times).
Under the medoid readout, the skeleton branch needs 0.359\,mJ and the action branch 7.12\,mJ.
The final RoboTwin 2.0 policy reads out with the medoid chunk of 10 steps.
It needs 10.7\,mJ per plan, and 90.2\,mJ with full computation.

\begin{figure*}[t]
  \renewcommand{\thefigure}{H.\arabic{figure}}%
  \centering
  \includegraphics[width=\textwidth]{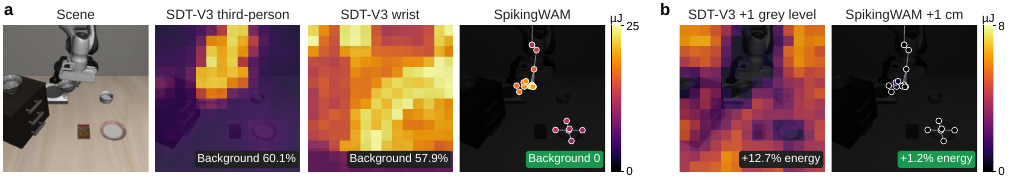}
  \caption{\textbf{Where events fall.}
  (a) Incremental events of one plan.
  The cells of SDT-V3 and the nodes of SpikingWAM share one colour scale.
  Labels give the share of events on the background over all 446 frames.
  Action tokens have no spatial position and are not drawn.
  (b) Events added by input noise of 1 grey level per camera frame or 1\,cm per object keypoint, shown on the third-person camera.
  Labels give the energy increase.}
  \label{fig:events}
\end{figure*}

\section{Evaluation Protocol}
\label{app:protocol}

Each of the four LIBERO suites has 10 tasks.
Each task runs one episode on each of its 50 official initial states, which gives 2,000 episodes.
The seven perturbations of LIBERO-Plus are camera, robot initial state, language, lighting, background, sensor noise and object layout, and each variant runs one episode.
RoboTwin 2.0 runs 100 seeds per task on both Easy and Hard, and none of the evaluation instructions appear in training.
The policy uses the EMA weights at the end of training.

\paragraph{Paired episodes.}
Two models are compared episode by episode on the same initial states or seeds.
We write the result as $a\,{:}\,b$, where $a$ counts the episodes in which only the first model succeeds and $b$ those in which only the second succeeds.

\section{Comparison with SkeleWAM}
\label{app:skelewam}

The main results of SkeleWAM~\cite{sheng2026skelewam} take object keypoints from the simulator state (\cref{sec:setup}).
We therefore compare it with SpikingWAM \S, whose object keypoints also come from the simulator state (\cref{tab:skelewam}).
The two have similar parameter counts.
SpikingWAM \S{} is 2.71 points higher in total.
The gain comes mainly from robot initial state and layout, by 13.3 and 17.8 points, and SpikingWAM \S{} is 11.4 points lower on language.
With the perception front end, SpikingWAM reaches 88.76\% on LIBERO-Plus (\cref{tab:main}), still above SkeleWAM \S.

\begin{table}[h]
  \centering
  \caption{\textbf{Success rates (\%) on LIBERO-Plus with object keypoints from the simulator state.}
  SkeleWAM results come from its paper.
  Both policies have 0.051B parameters.}
  \label{tab:skelewam}
  \footnotesize
  \setlength{\tabcolsep}{6pt}
  \begin{tabular}{@{}lc>{\columncolor{ourrow}[\tabcolsep][0pt]}c@{}}
    \toprule
    & SkeleWAM \S~\cite{sheng2026skelewam} & \multicolumn{1}{c@{}}{\textbf{SpikingWAM \S}} \\
    \midrule
    Camera & \textbf{96.1} & \textbf{96.1} \\
    Robot initial state & 75.2 & \textbf{88.5} \\
    Language & \textbf{89.0} & 77.6 \\
    Light & \textbf{96.6} & 95.3 \\
    Background & \textbf{96.7} & 95.8 \\
    Noise & \textbf{95.4} & 95.3 \\
    Layout & 69.0 & \textbf{86.8} \\
    Total & 87.7 & \textbf{90.41} \\
    \bottomrule
  \end{tabular}
\end{table}

\section{Statistics of Events and Attention}
\label{app:events}

We take the first demonstration of each of the 40 LIBERO tasks and keep one frame every 16 steps, which gives 446 frames.
Both cameras are rendered at $256\times256$ together with instance segmentation.
Pixels are split into four regions, namely background (table, floor and walls), arm, task objects (the objects of interest in the BDDL file) and other objects.
By pixel area, the background covers 64.3\%, the arm 13.6\%, the task objects 12.6\% and the other objects 9.6\%.
SDT-V3~\cite{yao2025scaling} uses its official ImageNet weights, with both images resized to $224\times224$.
The events at each position are split among the regions by the pixel shares inside its cell.
For pixel noise, each camera frame gets independent Gaussian noise with standard deviation $s/255$ before resizing.

Between consecutive plans, incremental firing keeps 48.1\% of the full events for SDT-V3 and 49.6\% for the skeleton branch of the SpikingWAM policy.
The two inputs therefore differ mainly in where the events fall (\cref{fig:events}).

The events of the perception front end are split among the regions in the same way.
The keypoint queries and the readout have no image position, so their 0.7\% of the events is not assigned to any region in \cref{fig:energy_regions}.

\paragraph{Attention.}
On the same 446 frames, with keypoints estimated by the perception front end, 75\% of the attention that the action tokens of SpikingWAM pay to nodes falls on the task objects.
Task object nodes make up only 57\% of all nodes, so the attention concentrates on the task objects.
For the image VLA TurboVLA~\cite{xie2026turbovla}, we use its public weights trained on LIBERO.
The attention of its action head follows the area of each region.
The background receives 0.62 of this attention, close to its 0.64 share of the area.

\section{Real Robot Experiments}
\label{app:real}

\paragraph{Platform and task.}
The experiments run on an SO-101 arm.
The table holds four cubes of different colours and two transparent cups, and the instruction is ``Put the green cube into the left cup.''
All three models use the same 324 teleoperated demonstrations.
X-VLA~\cite{zheng2025xvla} and $\pi_{0.5}$~\cite{intelligence2025pi05} are fine-tuned from their public weights.
SpikingWAM is trained from scratch with the same policy architecture as in simulation, and it reads out with the medoid chunk of 5 steps ($k=3$).

\paragraph{Keypoints.}
The perception front end (0.81M parameters) estimates the cube nodes at every plan.
A fine-tuned SDT-V3~\cite{yao2025scaling} estimates the cup nodes once at the start of the episode.
The label of a cup node is the gripper position at release in the demonstrations.
On 18 held out demonstrations, the median error of the cube nodes at the grasp time is 0.32\,cm.

\paragraph{Measuring cost.}
The energy of SpikingWAM is counted with the rules of \cref{app:energy} over all plans of the 10 trials, including the perception front end.
The energy of locating the cups and of reading the scene at the start of an episode is spread over the plans.
The energy of X-VLA and $\pi_{0.5}$ is estimated with the rules of \cref{sec:setup}, and image encoding counts the image streams that each model actually processes.
All three models run on the same H100 in BF16 with batch size 1, and X-VLA and $\pi_{0.5}$ use their official inference code.
Inference time is the mean time of one inference over 18 validation demonstrations, including preprocessing and excluding network transfer.
Peak memory covers loading and all inference.

\section{Role of Future Prediction}
\label{app:future}

The model that learns only actions removes the future skeleton tokens and their loss, and all other settings match SpikingWAM.
Both read out with PSR and generate no future skeleton at inference.
\Cref{tab:future} gives the results.

The two are on par on LIBERO.
SpikingWAM against the model that learns only actions gives 36\,:\,31 paired episodes on LIBERO and 607\,:\,451 on LIBERO-Plus (notation in \cref{app:protocol}).
The gap concentrates on the language and robot initial state perturbations, and the other five categories differ by at most 1.5 points.
On language, the gap concentrates on LIBERO-Spatial (85.6 against 74.4) and LIBERO-Goal (66.8 against 58.0).
Both suites need the instruction to pick out the target within the same scene.

\begin{table}[h]
  \centering
  \caption{\textbf{Removing future prediction (\%).}
  Object keypoints come from the perception front end, and both models read out with PSR.}
  \label{tab:future}
  \footnotesize
  \setlength{\tabcolsep}{6pt}
  \begin{tabular}{@{}lc>{\columncolor{ourrow}[\tabcolsep][0pt]}c@{}}
    \toprule
    & Actions only & \multicolumn{1}{c@{}}{\textbf{SpikingWAM}} \\
    \midrule
    LIBERO & 98.00 & \textbf{98.25} \\
    LIBERO-Plus & 87.21 & \textbf{88.76} \\
    \quad Camera & 94.4 & \textbf{94.6} \\
    \quad Robot initial state & 83.3 & \textbf{86.5} \\
    \quad Language & 69.7 & \textbf{76.1} \\
    \quad Light & 93.7 & \textbf{95.2} \\
    \quad Background & \textbf{93.8} & 92.9 \\
    \quad Noise & 94.0 & \textbf{94.3} \\
    \quad Layout & \textbf{84.7} & 84.1 \\
    \bottomrule
  \end{tabular}
\end{table}

\end{document}